\def\year{2019}\relax
\documentclass[letterpaper]{article} %DO NOT CHANGE THIS
\usepackage{aaai19}  %Required
\usepackage{times}  %Required
\usepackage{helvet}  %Required
\usepackage{courier}  %Required
\usepackage{url}  %Required
\usepackage{graphicx}  %Required
\usepackage{amsmath}
\usepackage{algorithmic}
\usepackage{array}
\usepackage[tight,footnotesize]{subfigure}
\begin{document}
\title{Renormalized Normalized Maximum Likelihood and Three-Part Code Criteria For Learning Gaussian Networks}
\author{Borzou Alipourfard \and Jean X. Gao\\
Department of Computer Science and Engineering\\
University of Texas at Arlington\\
Arlington, Texas 76019\\
}

\maketitle
\begin{abstract}
Score based learning (SBL) is a promising approach for learning Bayesian networks in the discrete domain. However, when employing SBL in the continuous domain, one is either forced to move the problem to the discrete domain or use metrics such as BIC/AIC, and these approaches are often lacking. Discretization can have an undesired impact on the accuracy of the results, and BIC/AIC can fall short of achieving the desired accuracy.  In this paper, we introduce two new scoring metrics for scoring Bayesian networks in the continuous domain: the three-part minimum description length  and the renormalized normalized maximum likelihood  metric.  We rely on the minimum description length principle in formulating these metrics. The metrics proposed are free of hyperparameters, decomposable, and are asymptotically consistent. We evaluate our solution by studying the convergence rate of the learned graph to the generating network and, also, the structural hamming distance of the learned graph to the generating network.  Our evaluations show that the proposed metrics outperform their competitors, the BIC/AIC metrics. Furthermore, using the proposed RNML metric, SBL will have the fastest rate of convergence with the smallest structural hamming distance to the generating network.
\end{abstract}

\section{Introduction}
\noindent A Bayesian network (BN) over a set of variables is a probabilistic graphical model where the dependencies between the variables are represented through a collection of edges among them in a directed acyclic graph (DAG)\cite{pearl2014probabilistic}. BNs have found extensive applications in diverse areas of engineering such as bioinformatics, image processing, and decision systems \cite{cowell2006probabilistic,spirtes2000constructing,friedman2004inferring}. 

In simple cases, experts can design BNs using their domain knowledge. However, in most applications, this approach is impractical.  It is, therefore, important to be able to learn and estimate a BN from observational data. There are two general approaches for learning a BN from data: \textit{constraint-based learning} \cite{spirtes2000causation} and \textit{score based learning } \cite{heckerman1995learning}. 

Score based learning has proved to be a promising approach for learning Bayesian networks \cite{chickering2002optimal,scutari2009learning,teyssier2012ordering,silander2012simple}. In score based learning, the learning of a Bayesian network from data is viewed as an optimization task. The optimization task consists of finding the network with the highest score where the candidate networks are scored with respect to the observations using a statistically suitable scoring metric.  Understandably, the choice of scoring metric plays an important role in the success of score based learning algorithms.

For Bayesian networks on discrete domains, various scoring metrics have been proposed, each formulated based on different set of statistical assumptions and motivations \cite{heckerman1995learning,silander2008factorized,bouckaert1993probabilistic}. For Bayesian networks over continuous domains, however, there are very few scoring criteria. Therefore, to employ score based learning in the continuous domain, one is most often either forced to use the BIC/AIC metrics or to convert the variables into discrete counterparts. These approaches, however, are often lacking. 

The most common discretization method used in practice is to heuristically partition the range of the continuous features into several mutually exclusive and exhaustive regions. Doubtless, the choice of discretization policy can have a significant undesired impact on the accuracy of the results \cite{daly2011learning}.  To minimize the unfavorable effects of discretization, Friedman et~al. proposed a more principled discretization policy where discretization of a continuous feature is based on its relation to other variables in the network \cite{friedman1996discretizing}. However, the scoring metric resulting from this discretization policy is not decomposable. Lack of decomposability makes the search for the highest scoring network computationally challenging as most search algorithms require the decomposability of the scoring metric.

In the absence of discretization, there are only three scoring metrics that are directly applicable to BNs on domains containing continuous variables; the AIC, BIC and the BGe score \cite{akaike1998information,schwarz1978estimating,geiger1994learning}.  The BIC/AIC metrics can fall short of achieving the desired accuracy. The BGe score lacks a closed form expression and requires specification of a set of hyperparameters which demands extensive domain knowledge or the use of a portion of the data.  

In this paper, we describe two novel scoring metrics for Bayesian networks in the continuous domain based on the minimum description length principle (MDL).  Our work draws inspiration from the use of the MDL principle in problems of variable selection in regression \cite{barron1998minimum,hansen2001model}. Thus we expect it to carry the advantages that MDL offers versus the AIC and BIC in the problems of variable selection in regression to the problem of finding a suitable Bayesian network \cite{grunwald2007minimum}; learning a Bayesian network can be thought of as selecting predictors for a set of variables where the predictors are constrained to follow a particular order. Both scoring metrics proposed here are free of hyperparameters, decomposable, and are asymptotically consistent.
 
In the next section, we formally introduce the Bayesian Gaussian networks. We then formulate a crude three-part code scoring metric for Bayesian networks in part 3 of our paper. In section 4, we propose a renormalized normalized maximum likelihood scoring scheme for continous domain Bayesian networks under the assumption that the model class under consideration consists of only Bayesian Gaussian networks. Afterward, we study the asymptotic properties of the two proposed scoring metrics in section 5.  We evaluate and compare the performance of the proposed scoring metrics to the BIC and the AIC metrics using simulated data in section 6. Our results suggest that the scoring metrics proposed here consistently perform better than BIC and AIC for both sparse and dense graphs.   

\section{Gaussian Networks}
Throughout this paper, we consider a domain, $X_m=\{x_1, x_2, ..., x_m\}$ of $m$ continuous variables. Let $\rho(\vec{x})$ denote the joint probability density function over $X_m$. A Bayesian network over $X_m$ is a pair $B = (B_P, B_S)$, where $B_S$ encodes a set of conditional independence statements and $B_P$ is a set of local conditional probabilities associated with $B_S$. In particular, $B_S$ is a collection of sets $\{\pi_i | \pi_i \subseteq \{x_1, x_2, ...,x_{i-1}\}, i = 1,2, ...,m \}$ such that $\pi_i$ renders $x_i$ and $\{x_1, x_2, ...,x_{i-1}\}$ conditionally independent:
\begin{equation} \label{eq0}
\rho(\vec{x}) = \prod_{i=1} ^{m}{\rho(x_i|\pi_i )}.
\end{equation}
Assuming that the joint probability distribution function of $\vec{x}$ is a multivariate normal distribution, we can write:
\begin{small}
\begin{equation}
\rho(\vec{x}) = \eta(\vec{\mu}, \Sigma^{-1}) = (2\pi)^{-m/2}|\Sigma|^{-1/2}e^{(\vec{x}-\vec{\mu})^{'}\Sigma^{-1}(\vec{x}-\vec{\mu})},
\end{equation}
\end{small}where $\vec{\mu}$ is the m-dimensional mean vector, $\Sigma$ is the $m\times{m}$ covariance matrix, $|.|$ is the determinant operation, and $(.)^{'}$ is the transpose operation.  This distribution can be factorized into the product of $m$ conditional distributions:
\begin{small}
\begin{equation}
\begin{aligned}
\rho(\vec{x}) {} & = \prod_{i=1}^{i=m}\rho(x_i|x_1, ..., x_{i-1})  \\
 & = \prod_{i=1}^{m} \eta(\mu_{i} + \sum_{j=1}^{i-1} b_{ij}(x_j - \mu_j), 1/{\tau_i}),
\end{aligned}
\end{equation}
\end{small}where $\tau_i$ is the residual variance of the node $x_i$, $\mu_i$ is the unconditional mean of $x_i$, and $b_{ij}$ is a measure of the extent of partial correlation between nodes $x_i$ and $x_j$ . Such a distribution corresponds to a Bayesian network $(B_P, B_S)$ if:
\begin{small}
\begin{equation}
\begin{aligned}
\forall i: \rho(x_i|x_1, ..., x_{i-1}) {} & = \rho(x_i|\pi_i) \\
& = \eta(\mu_{i} + \sum_{x_j \in \pi_i} b_{ij}(x_j - \mu_j), 1/\tau_i ),
\end{aligned}
\end{equation}
\end{small}where $\pi_i$'s correspond to the parent sets specified in $B_s$. In other words, a multivariate Gaussian distribution corresponds to a BN, $(B_P, B_S)$, if $ \forall b_{ij}\neq 0: x_j \in \pi_i $ \cite{shachter1989gaussian}. The Bayesian network is minimal when there is an arc from $x_j$ to $x_i$ if and only if $b_{ij}\neq 0$. This network is also referred to as the I-map of the  probability distribution. 

Instead of the above parametrization of a multivariate Normal distribution, we opt to work with the following parametrization:
\begin{small}
\begin{equation}
\begin{aligned}
\forall i: \rho(x_i|x_0 = 1, \pi_i) =  \eta( \sum_{x_j \in \pi_i } \beta_{ij}x_j + \beta_{i0}, 1/\tau_i ).
\end{aligned}
\end{equation}
\end{small}Every instantiation of parameters of one model corresponds to an instantiation of parameters of the second model; note that the new parametrization corresponds to a network with $m+1$ nodes with $x_0 = 1$, and the unconditional means of all other nodes set to zero. Thus a Bayesian network, $B = (B_P, B_S)$,  can be parametrized by $\{(\vec{\beta_{i}},\tau_i)| i =1, 2, ..., m\}$. We call such a Bayesian network, a Gaussian network; if further, the network is minimal, we call it a minimal Gaussian network. In the rest of the paper, for notational convenience, we represent the set $\{ \pi_i, x_0\}$ simply by $\pi_i$. We also write $k_i$ for the cardinality of the $\pi_i$, $|\pi_i|$, and $Pa_i$ for the values of the variables $\pi_i$.

\section{A Crude Three-Part Code Scoring Metric}
In this section, we propose our first scoring metric for Bayesian networks based on the MDL principle. We assume that the reader is familiar with the basics of MDL principle \cite{jorma1998stochastic,rissanen1986stochastic}. In short, the MDL principle states that the model most suitable for a given set of observations is the one that encodes the observations using the fewest symbols.  

We wish to measure how well a Bayesian network structure, $B_S$ fits the observed data. Motivated by the MDL principle, we can alternatively evaluate how compact a description a Bayesian network structure can provide for the observations. We propose two methods for encoding observations using Gaussian networks. The first method proposed in this section is a crude and a heuristic encoding scheme. In the next section of the paper, we optimize this coding scheme to formulate a min-max optimal encoding of the observations known as normalized maximum likelihood coding \cite{grunwald2007minimum}. 

The three-part code encodes the observations in three steps. The first part of the code describes the Bayesian network structure and the second and the third part encode the observations using the network structure coded in the first part. The total description length then serves as a measure of how well the network fits the observations. 

In this three-part code, we extend the MDL formulations of Lam et~al. for Bayesian networks over discrete domains to continuous domains \cite{lam1994learning}. First, to encode the structure of the network, we simply enumerate the parents of each node. For a node with $k_i$ parents we use $k_i\ln{m}$ nats to list its parents. Therefore, the encoding length of the structure of the network can be written as: 

\begin{small}
\begin{equation}
\begin{aligned}
L_1 = L(B_S) = \sum_{i=1}^{m} k_i\ln(m).
\end{aligned}
\end{equation}
\end{small}Having coded the network structure, we now describe how we encode the observations in the remainder of the code. Assume that we have observed the data $y^n = [x_1^n, ..., x_m^n]$ of sample size $n$. Our coding scheme is to first encode the values of the root nodes (nodes without parents) and then to encode the values of nodes whose parent values has already been encoded. We continue this process iteratively, descending the Bayesian network until we reach the leaves of the network.  

Now, suppose that we have already encoded the values for $\{x_1^n, ..., x_{i-1}^n\}$ and we wish to encode the values of $x_i$. Since we have assumed that data is sampled from a multivariate normal distribution, we have:
\begin{small}
\begin{equation}
\begin{aligned}
\rho(x_i^n|x_1^n, ..., x_{i-1}^n) = (2\pi\tau_i)^{-n/2}e^{\dfrac{-||x_i^n - Pa_i^n \vec{\beta_i} ||^2}{2\tau_i} },
\end{aligned}
\end{equation}
\end{small}where $Pa_i^n$ is the $n\times{k_i}$ matrix of the values of the parents of $x_i$. Within this framework, the problem of optimal encoding of $x_i^n$ and the parameters $(\vec{\beta_i}, \tau_i)$ is equivalent to the problem of encoding a response variable given the values of predictor variables in linear regression. Here, the response variable is $x_i$, and the predictor variables are $\pi_i$. Following works of Rissanen, in such a setting, the shortest code for encoding the values of $x_i$ has a length of \cite{rissanen1983universal,jorma1998stochastic}:
\begin{small}
\begin{equation}
\begin{aligned}
L_2 = L(x_i^n|x_1^n, ..., x_{i-1}^n) {} & = L_{21} + L{22} \\
& = \frac{k_i}{2} \ln{n}  -\ln{\rho(x_i^n|Pa_i^n; \hat{\vec{\beta_i}}, \hat{\tau_i})} \\
& = \frac{k_i}{2} \ln{n} + \frac{n}{2}\ln{(2\pi e \hat{\tau_i})},					
\end{aligned}
\end{equation}
\end{small}
where $\hat{\tau_i}$ and $\hat{\vec{\beta_i}}$ are the maximum likelihood estimates (MLE) of $\tau_i$ and $\vec{\beta_i}$. More specifically, we encode the values of $x_i^n$ in two parts; in the first part, we encode the MLE parameters  $\hat{\vec{\beta_i}}$, and in the second part we encode the values of $x_i^n$ using the distribution $\rho(x_i^n|\pi_i^n; \hat{\vec{\beta_i}}, \hat{\tau_i})$. Such a coding scheme is referred to as a crude two part coding in the MDL literature \cite{grunwald2007minimum}. 

Thus the total description length of the observed data will be:
\begin{small}
\begin{equation} \label{eq9}
\begin{aligned}
L(x_1, ..., x_m|x_0)	= & \frac{n}{2} \sum_{i=1}^{m} \ln{(2\pi e\hat{\tau_i})} + \\ & [\ln(n)/2 + \ln(m)] \sum_{i=1}^{m}k_i .
\end{aligned}
\end{equation}
\end{small}It is insightful to compare this coding metric to BIC for Gaussian networks and the MDL metric proposed for BNs over discrete domains. Comparing the penalty terms in the proposed MDL scoring metric for Gaussian networks with the MDL scoring metric of discrete networks, one observes that the penalty term for discrete networks is exponential in the number of parents while it only grows linearly for Gaussian networks. The reason is that the dimensionality of the parameter space for discrete Bayesian networks increases exponentially with an increasing number of parents while the parameter space of multivariate Gaussian distribution is polynomial in the number of parents. The proposed MDL metric is essentially the same as the BIC metric with the addition of the penalty term $\sum_{i=1}^{m} k_i\ln(m)$ which accounts for the network structure, $B_S$. 
\subsection{Inefficieny of Two Part Codes}
The coding scheme introduced above is not Kraft-tight \cite{rissanen1986stochastic}. In particular, consider the code proposed for encoding values of  $x_i^n$  given values of $\{x_1^n, ..., x_{i-1}^n\}$:
\begin{small}
\begin{equation}
\begin{aligned}
L(x_i^n|x_1^n, ..., x_{i-1}^n) {} & = L_{21} + L_{22} \\
& = \frac{k_i}{2} \ln{n} + \frac{n}{2}\ln{(2\pi e \hat{\tau_i})}		.			
\end{aligned}
\end{equation}
\end{small}
Note that once we decode $L_{21}$ ($L_{21}$ contains information on the MLE values for the regression coefficients \boldmath$({\hat{\vec{\beta_i}}})$ \unboldmath) the set of possible values for $x_i^n$ become restricted to those for which $\hat{\vec{\beta_i}}(x_i^n) =$\boldmath$({\hat{\vec{\beta_i}}})$ \unboldmath. Therefore, this coding scheme is inefficient and the data can be coded using fewer bits. Normalized maximum likelihood (NML) codes are a variation of the two-part coding scheme where this inefficiency of the crude two-part coding is addressed \cite{rissanen1986stochastic,rissanen2000mdl}. 

\section{Normalized Maximum Likelihood}
The normalized maximum likelihood distribution with respect to a class of probability distributions parametrized by a $K$ dimensional parameter vector $\theta$, $C_\theta(x) = \{P(x;\theta)| \theta \in R^K\}$, is defined as:
\begin{small}
\begin{equation}
\begin{aligned}
P_{nml}(x) = \dfrac{P(x; \hat{\theta}(x))}{\int P(y; \hat{\theta}(y)) dy},
\end{aligned}
\end{equation}
\end{small}where 
\begin{small}
\begin{equation}
\begin{aligned}
\hat{\theta}(x) = \underset{\theta}{\scriptsize{argmax}} \left[P(x;\theta)\right], P(x;\theta) \in C_\theta(x).
\end{aligned}
\end{equation}
\end{small}In normalized maximum likelihood codes, instead of using a three-part code, we encode each observation with a single code using the NML distribution:
\begin{small}
\begin{equation}
\begin{aligned}
L_{nml}(x) = -\ln{(P_{nml}(x))}.
\end{aligned}
\end{equation}
\end{small}We are now ready to formulate the NML pdf for a Gaussian network. 

A Gaussian network structure over m variables defines a class of probability distributions parametrized by $\theta = \{ (\vec{\beta_i}, \tau_i)|i=1, ..., m\}$:
\begin{small}
\begin{equation}
\begin{aligned}
{} & G_{\theta}(y^n) = \{\rho(y^n; \theta)| \theta \in R^{m+\sum_{i=1}^{m} k_i }\}, \\
& \rho(y^n; \theta) = \rho(x_1^n, ...,x_m^n; \theta) = \prod_{i=1}^{m} \rho(x_i^n|\pi_i^n; \vec{\beta_i}, \tau_i), \\
\end{aligned}
\end{equation}
\end{small}where:
\begin{small}
\begin{equation}
\begin{aligned}
 \rho(x_i^n|\pi_i^n; \vec{\beta_i}, \tau_i) {} & = \eta(\vec{\beta_i} Pa_i^n, 1/\tau_i) \\
 & = (\dfrac{1}{2\pi \tau_i})^{n/2} \exp(\dfrac{1}{2\tau_i} || x_i^n - Pa_i^n \vec{\beta_i} ||^2).
\end{aligned}
\end{equation}
\end{small}Let $\hat{\theta}(y^n)= \{ (\hat{\vec{\beta_i}}(y^n), \hat{\tau_i})|i=1, ..., m \}$ denote the MLE estimates of $\vec{\beta_i}$ and $\tau_i$:
\begin{small}
\begin{equation}
\begin{aligned}
{} & \hat{\vec{\beta_i}}(y^n)  = \hat{\vec{\beta_i}}(x_i^n, Pa_i^n)= (n\Sigma_i)^{-1}{Pa_i^n}^{'} x_i^n, \\
& \Sigma_i = n^{-1} Pa_i^{'}Pa_i, \\
& \hat{\tau_i}(y^n) = \hat{\tau_i}(x_i^n, Pa_i^n) = 1/n || x_i^n - Pa_i^n \hat{\vec{\beta_i}} ||^2, \\
& \rho(x_i^n|Pa_i^n; \hat{\vec{\beta_i}}(y^n), \hat{\tau_i}(y^n)) =  (2 \pi e \hat{\tau_i}(y^n) ) ^ {-n/2}.
\end{aligned}
\end{equation}
\end{small}The integral in the denominator of the NML distribution does not exist for $G_{\theta}(y^n)$ \cite{miyaguchi2017normalized}. We write down the constrained NML density as below: 
\begin{small}
\begin{equation} \label{eq:16}
\begin{aligned}
P_{nml}(x; \theta^0) = \dfrac{P(x; \hat{\theta}(x))}{\int_{Y(\theta^0)} P(y; \hat{\theta}(y)) dy},
\end{aligned}
\end{equation}
\end{small}the constrained NML density is only defined for $y^n \in Y(\theta^0)$:
\begin{small}
\begin{equation}
\begin{aligned}
Y(\theta^0) = \{y^n| \hat{\theta}(y^n) \in \theta^{0} \}.
\end{aligned}
\end{equation}
\end{small}In the case of $G_{\theta}(y^n)$, we can specify $\theta^{0}$ using the following set of hyperparameters:
\begin{small}
\begin{equation}
\begin{aligned}
{} & \theta^{0} = (\tau^0, R^0), \\
    & \tau^0 = \{\tau_i^0| i =1, ..., m\}, \\
    & R^0 = \{R_i^0| i =1, ..., m\}.
\end{aligned}
\end{equation}
\end{small}Let these hyperparameters define the $Y(\theta^0)$ as below:
\begin{small}
\begin{equation}
\begin{aligned}
Y(\theta^0) {} &  = Y(\tau^0, R^0) {} & \\ 
& = \{y^n: x_1^n \in X_1(\tau_1^0, R_1^0), ... \\ 
& ,x_m^n \in X_m(\tau_m^0, R_m^0, x_1^n, ..., x_{m-1}^n) \}.
\end{aligned}
\end{equation}
\end{small}and
\begin{small}
\begin{equation}
\begin{aligned}
 X_i(\tau_i^0, R_i^0, x_1^n, ..., x_{i-1}^n) = {} & \{x_i^n| \hat{\tau_i}(x_i^n, Pa_i^n) \geq \tau_i^0, \\
& \hat{\vec{\beta_i}}^{'}(x_i^n, Pa_i^n) \Sigma_i \hat{\vec{\beta_i}}(x_i^n, Pa_i^n) \leq R_i^0\}, \\
\end{aligned}
\end{equation}
\end{small}The numerator of \ref{eq:16} can be easily calculated as:\\
\begin{small}
\begin{equation}
\begin{aligned}
\rho(y^n; \hat{\theta}(y^n)) {} &= \prod_{i=1}^m \rho(x_i^n|Pa_i^n; \hat{\tau}(x_i^n, Pa_i^n), \hat{\vec{\beta_i}}(x_i^n, Pa_i^n)) \\
& = \prod_{i=1}^m (2\pi e\hat{\tau}(x_i^n, Pa_i^n))^{-n/2} .
\end{aligned}
\end{equation}
\end{small}We can calculate the denominator by first writing down the factored form of the density:
\begin{small}
\begin{equation}
\begin{aligned}
{} & \int_{Y(\theta^0)} \rho(y^n; \hat{\theta}(y^n)) dy^n  = \int_{X_1(\tau_1^0, R_1^0)} ... \\
& \int_{X_m(\tau_m^0, R_m^0, Pa_i^n)} \prod_{i=1}^m \rho(x_i^n|Pa_i^n; \hat{\vec{\beta_i}}(y^n), \hat{\tau_i}(y^n)) dx_1^n ... dx_m^n .
\end{aligned}
\end{equation}
\end{small}Since only the factor $\rho(x_m^n|Pa_m^n; \hat{\vec{\beta_m}}(y^n), \hat{\tau_m}(y^n))$ is a function of $x_n^m$, we can take the other factors out of the last integral. We now write down this last integral for a given value of $\{x_1^n, ..., x_{m-1}^n\}$:
\begin{small}
\begin{equation}
\begin{aligned}
\int_{X_m(\tau_m^0, R_m^0, Pa_i^n)} \rho(x_m^n|Pa_i^n; \hat{\vec{\beta_i}}(y^n), \hat{\tau_i}(y^n)) dx_m^n .
\end{aligned}
\end{equation}
\end{small}Note that both the region of integration and the MLE estimates are functions of $\{x_1^n, ..., x_{m-1}^n\}$. Using sufficient statistics, the value of this integral was calculated in \cite{roos2004mdl} :
\begin{small}
\begin{equation}
\begin{aligned}
C^m(\tau^0, R^0) {} & =\int_{X_m(\tau_m^0, R_m^0, Pa_i^n)} \rho(x_m^n|Pa_i^n; \hat{\vec{\beta_i}}(y^n), \hat{\tau_i}(y^n)) dx_m^n \\
& = \dfrac{4n^{n/2} (\dfrac{R_m^0}{\tau_m^0})^{-k_m/2}}{(2e)^{n/2}k_m^2 \Gamma{(n-k_m)} \Gamma{(k_m/2)}} .
\end{aligned}
\end{equation}
\end{small}Interestingly, the above factor is independent of the values of $\{x_1^n, ..., x_{m-1}^n\}$. As we will see later, this independence will make way for a scoring metric that is decomposable and local to the nodes of the graph. \\
The denominator of the constrained NML distribution is then:
\begin{small}
\begin{equation}
\begin{aligned}
C(\tau^0, R^0) {} & =\prod_{i=1}^m \dfrac{4n^{n/2} (\dfrac{R_i^0}{\tau_i^0})^{-k_i/2}}{(2e)^{n/2}k_i^2 \Gamma{(n-k_i)} \Gamma{(k_i/2)}}.
\end{aligned}
\end{equation}
\end{small}In the expression above, due to the terms $(\dfrac{R_i^0}{\tau_i^0})^{-k_i/2}$, the hyperparameters $(\tau^0, R^0)$  have different effects on the score of different networks structures. To get rid of such effects, Rissanen proposes a second level normalization \cite{rissanen2000mdl}.
\subsection{Renormalized Normalized Maximum Likelihood Scoring Metric}
Let $\hat{\tau}^0(y^n) = \{\hat{\tau}_i^0(y^n)|i=1, 2, ...,m\}$ and $\hat{R}^0(y^n) = \{\hat{R}_i^0(y^n)|i=1, 2, ...,m\}$ denote the MLE estimates of $\tau^0$ and $R^0$ :
\begin{small}
\begin{equation}
\begin{aligned}
& \hat{\tau}_i^0(y^n)= \hat{\tau}_i(x_i^n, Pa_i^n) = 1/n || x_i^n - Pa_i^n \hat{\vec{\beta_i}} ||^2, \\
& \hat{R}_i^0(y^n) = \hat{R}_i(x_i^n, Pa_i^n) = \hat{\vec{\beta_i}}(x_i^n, Pa_i^n)^{'} \Sigma_i \hat{\vec{\beta_i}}(x_i^n, Pa_i^n),
\end{aligned}
\end{equation}
\end{small}where
\begin{small}
\begin{equation}
\begin{aligned}
\hat{\vec{\beta_i}}(x_i^n, Pa_i^n) =(n\Sigma_i)^{-1}{Pa_i^n}^{'} x_i^n. 
\end{aligned}
\end{equation}
\end{small}The renormalized NML (RNML) probability distribution is then given by:
\begin{small}
\begin{equation} \label{eq:28}
\begin{aligned} 
\bar{\rho}(y^n) = \dfrac{\rho_{nml}(y^n; \hat{\tau}^0(y^n), \hat{R}^0(y^n))}{\int_{Z(\tau^1, \tau^2, R^1, R^2)} \rho_{nml}(z^n; \hat{\tau}^0(z^n), \hat{R}^0(z^n)) dz^n},
\end{aligned}
\end{equation}
\end{small}where the region of integration is given by the hyperparameters:
\begin{small}
\begin{equation}
\begin{aligned}
& \tau^1 = \{\tau_i^1|i=1, 2, ..., m\} \\
& \tau^2 = \{\tau_i^2|i=1, 2, ..., m\} \\
& R^1 = \{R_i^1|i=1, 2, ..., m\} \\
& R^2 = \{R_i^2|i=1, 2, ..., m\} \\
\end{aligned}
\end{equation}
\end{small}with $Z(\tau^1, \tau^2, R^1, R^2)$ defined as:  
\begin{scriptsize}
\begin{equation}
\begin{aligned}
Z(\tau^1, \tau^2, R^1, R^2) = \{z^n| \forall i=1, 2, ..., m : & \\
\tau_i^2 \geq \hat{\tau_i}^0(z^n) \geq \tau_i^1, R_i^1 \geq \hat{R}_i^0(z^n) \geq R_i^2\} 
\end{aligned}
\end{equation}
\end{scriptsize}Inserting the density for the NML distribution into (\ref{eq:28}), with the boundry conditions above, the RNML distribution can be calculated as:
\begin{scriptsize}
\begin{equation} \label{eq:31}
\begin{aligned} 
\bar{\rho}(y^n) = \prod_{i=1}^m \dfrac{(\hat{\tau}_i)^{-n/2} (n\pi)^{-n/2} \Gamma(k_i/2)\Gamma(n-k_i) (\dfrac{\hat{R}_i^0(y^n)}{\hat{\tau}_i^0(y^n)})^{-k_i/2}}{\ln(\dfrac{\tau_i^2}{\tau_i^1}\ln(\dfrac{R_i^2}{R_i^1}))},
\end{aligned}
\end{equation}
\end{scriptsize} where we have used the RNML calculations for Gaussian distribution from \cite{roos2004mdl}, together with the property of our parametriziation that allows the integrals to be calculated independent of each other in solving the RNML distribution. Dropping the terms independent of the network structure, the RNML code can be written as:
\begin{scriptsize}
\begin{equation} \label{eq:32}
\begin{aligned} 
{} &  L_{RNML}(y^n) = -\ln(\bar{\rho}(y^n)) \\
& = \sum_{i=1}^m \left(\dfrac{n}{2} \ln(\hat{\tau_i}(y^n)) -\ln(\Gamma(\dfrac{k_i}{2})) - \ln(\Gamma(\dfrac{n-k_i}{2})) + \dfrac{k_i}{2} \ln(\dfrac{\hat{R}_i^0(y^n)}{\hat{\tau_i}(y^n)}) \right). 
\end{aligned}
\end{equation}
\end{scriptsize}Using Stirling's approximation, we can simplify the above expression:
\begin{scriptsize}
\begin{equation} \label{eq:33}
\begin{aligned} 
{} &  L_{RNML}(y^n) = -\ln(\bar{\rho}(y^n)) \\
& = \sum_{i=1}^m \left((n-k_i)\ln(\dfrac{\hat{\tau_i}(y^n)}{n-k_i} ) + k_i\ln(\dfrac{\hat{R}_i^0(y^n)}{k_i})+\ln(k_i(n-k_i))\right)
\end{aligned}
\end{equation}
\end{scriptsize}
Equations (\ref{eq:33}) and (\ref{eq:32}) provide a closed-form expression for the scoring of a Gaussian Bayesian network based on the RNML metric. Note that both these expressions are free of hyperparameter and are decomposable.
\section{Asymptotic Behavior}
It is well known that BIC prefers minimal I-maps over other network structures for large sample sizes \cite{bouckaert1993probabilistic}. Examining the equation (\ref{eq9}), it is clear that the asymptotic behavior of the three-part coding metric is equivalent to that of the BIC metric. We now show that the RNML scoring metric also prefers networks that are minimal I-maps.

\textbf{Theorem 1.} Let $X_m$ be a set of variables, $T$ be an ordering on the variables in $X_m$ and $\rho$ be a probability distribution over $X_m$. Let $y^n$ be a sample generated from $\rho$. Let $B_s$ be a minimal I-map Bayesian network of $\rho$ and let $B_{s^\prime}$ be any other network structure. Furthermore, let both $B_{s^\prime}$ and $B_s$ be consistent with the ordering $T$. We have:
\begin{equation}
\begin{aligned} 
L_{RNML}(B_s, y^n) < L_{RNML}(B_{s^\prime}, y^n)
\end{aligned}
\end{equation}
That is the network corresponding to the minimal I-map has the lowest description length based on the RNML distribution. 

\textbf{Proof}. We consider two cases. In the first case, we assume that $B_{s^\prime}$  is a non-minimal I-map of $P$. In the second case, we consider a $B_{s^\prime}$ that is not an I-map of $P$.

Assuming that $B_{s^\prime}$ is a non-minimal I-map of $P$, then the variance of the residual, $\tau_i$, of each node is the same in both $B_{s^\prime}$ and $B_s$. Dropping the factors less than $O(n)$ we have:
\begin{scriptsize}
\begin{equation}
\begin{aligned} 
{} & L_{RNML}(B_s, y^n)  - L_{RNML}(B_{s^\prime},  y^n) \\
& = -\sum_{i=1}^{m}  -\ln(\Gamma(\dfrac{n-k_i^{s}}{2})) + \ln(\Gamma(\dfrac{n-k_i^{s^\prime}}{2})).
\end{aligned} 
\end{equation}
\end{scriptsize}
Since $B_{s^\prime}$ is a non-minimal I-map of $P$ we also have:
\begin{scriptsize}
\begin{equation}
\begin{aligned} 
\forall i: k_i^{_s} \leq k_i^{_{s^\prime}}.
\end{aligned} 
\end{equation}
\end{scriptsize}
Therefore, we have: 
\begin{scriptsize}
\begin{equation}
\begin{aligned} 
{} & L_{RNML}(B_s, y^n)  - L_{RNML}(B_{s^\prime},  y^n) \leq 0
\end{aligned} 
\end{equation}
\end{scriptsize}Now assume that $B_{s^\prime}$ is a non-minimal I-map of $P$. Thus there exists at least one node, $x_i$, such that $\pi_i^{_s} \not\subseteq \pi_i^{_{s^\prime}}$ or equivalently, $\exists x_j \in \pi_i^{_s}$ such that $x_j \not\in \pi_i^{_{s^\prime}}$. Without loss of generality, we assume that this consists of the only difference between the two networks. Now consider a network structure, $B_c$, that is exactly equal to network $B_s$ except for the parent set of the node $x_i$ where $\pi_i^{_c} = \pi_i^{_s} \cup \pi_i^{s^\prime}$. Therefore, $B_c$ is a non-minimal I-map of the probability distribution and from the previous result we know that $L_{RNML}(B_s, y^n) \leq L_{RNML}(B_c,  y^n)$ . We now show that $L_{RNML}(B_c,  y^n) \leq L_{RNML}(B_{s^\prime},  y^n)$. 

We can transform the network $B_c$ to the network $B_{s^\prime}$ by removing the nodes $\{x_k: x_k \in \pi_i^{_s}, x_k \not\in \pi_i^{s^\prime} \}$ from $\pi_i^{s^\prime}$. Doing so will increase the residual variance of the node $x_i$ by at least $\beta_{ik}^2 \hat{\tau_k}(y^n)$ . More specifically, using equation \ref{eq:33} dropping the node $x_k$ will increase the $L_{RNML}(B_c, y^n)$: 
\begin{scriptsize}
\begin{equation}
\begin{aligned} 
& \Delta L_{RNML}  = \\
& (n-k_i^c)ln(\dfrac{\hat{\tau}_i^c(y(n))}{n-k_i^c}) - (n-k_i^c-1)ln(\dfrac{\hat{\tau}_i^c(y(n)) - \beta_{ik}^2 \hat{\tau_k}(y^n)}{n-k_i^c-1}).
\end{aligned} 
\end{equation}
\end{scriptsize}Keeping only the terms of the factor $O(n)$, we can simplify the above expression:
\begin{scriptsize}
\begin{equation}
\begin{aligned} 
\Delta L_{RNML} {} & =  (n-k_i^c)ln\left(\dfrac{ \dfrac{\hat{\tau}_i^c(y(n))}{\hat{\tau}_i^c(y(n)) - \beta_{ik}^2 \hat{\tau}_k^c(y^n)} }{ \dfrac{n-k_i^c}{n-k_i^c-1} }\right) \\
& \stackrel{n\to\infty}{=} (n-k_i^c)ln\left( \dfrac{\hat{\tau}_i^c(y(n))}{\hat{\tau}_i^c(y(n)) - \beta_{ik}^2 \hat{\tau}_k^c(y^n)} \right) \\
& = O(n).
\end{aligned} 
\end{equation}
\end{scriptsize}Hence we can see that asymptotically as ${n\to\infty} $, $L_{RNML}(B_c,  y^n)$ is smaller than  $L_{RNML}(B_{s^\prime},  y^n)$  by a factor of $O(n)$. 

\section{Numerical Evaluation}
We evaluate and compare the performance of the proposed scoring metrics to the BIC and the AIC metrics using simulated data. We study the properties of these four metrics on two levels: a low dimensional setting where the number of the nodes of the graph is small, $m \in \{4, 5\}$, and lager graphs having $m \in \{8, 10, 15\}$ nodes. In low dimensional setting, the number of possible generating graphs is limited. It is, therefore, possible to evaluate the performance of these four metrics in detail since we can calculate the score of all the possible generating networks. In this setting, we examine how the true generating network structure is ranked among all other networks by the different scoring criteria.

With larger DAGs, the number of possible generating graphs is exponentially high and such evaluation of the performance of the scoring metrics is computationally challenging. Hence, we decided to compare the performance of these metrics by using a structural distance measure between the highest scoring networks (found using dynamic programming \cite{silander2012simple}) and the true generating network. 

We chose Structural Hamming Distance (SHD) as our measure of performance. SHD is a function of the number of edge addition/deletion or reversals required to convert one DAG to another \cite{tsamardinos2006max}. While such a structural distance measure can only provide a heuristic summary of the performance of these metrics, nevertheless, a comparison in terms of such structural errors has a desirable intuitive interpretation. Furthermore, SHD compares the similarity of BNs in a causal context \cite{tsamardinos2006max,de2009comparison}. Therefore, it also serves as a tool to examine the applicability of the proposed metrics in identifying the generating causal structure.
\begin{figure}
\centerline{
\begin{subfigure}[]{}
  \includegraphics[width=2.5in]{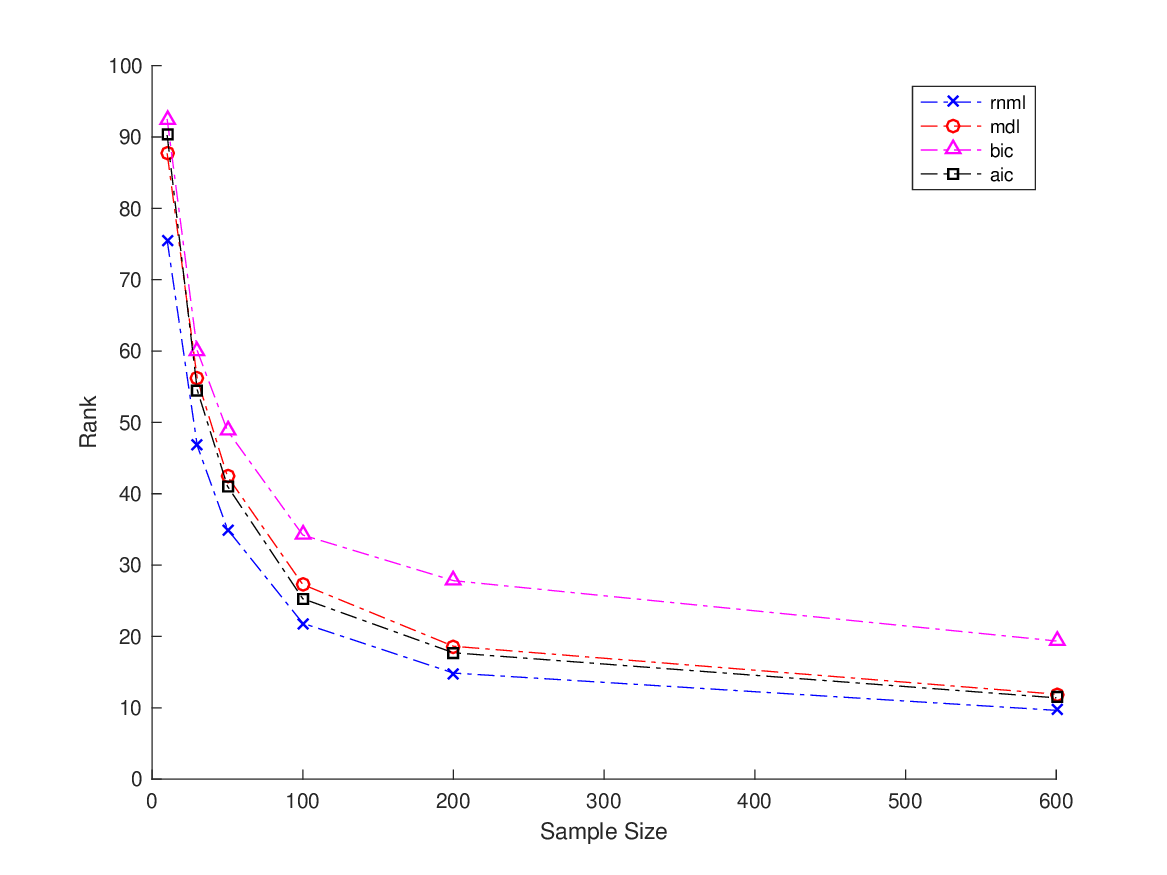}
\end{subfigure}}
\hfil
\begin{subfigure}[]{}
  \includegraphics[width=2.5in]{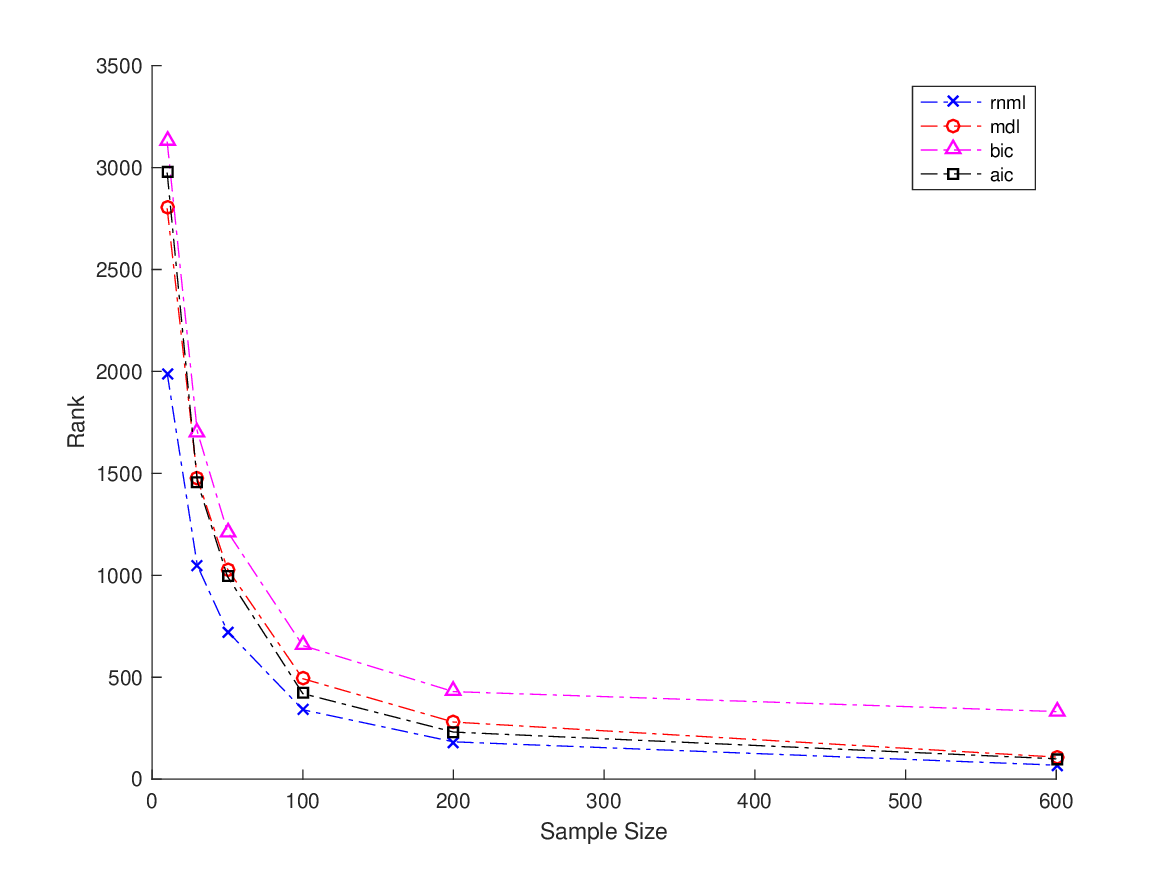}
\end{subfigure}
\caption{The average rank of the generating network structure for networks having $m$ nodes, $m \in \{4,5\}$, over 500  iterations plotted against the sample size. The upper plot shows the results for networks having four nodes, while the convergence rate for networks having five nodes is shown in the lower plot. }
\label{Fig1}
\end{figure}
\section{Performance in Low Dimensional Setting}
In this experiment, we compare how the generating network structure is ranked among all the possible networks by the different scoring criteria.

 First, a random DAG was chosen among all DAGs on $m$ nodes, $m \in \{4,5\}$. A sample was then recursively generated from the selected DAG  starting from the root node down to the leaves using the following equation:

\begin{small}
\begin{equation}
\begin{aligned}
x_i = \mu_i + \sum_{x_i \in \pi_i} b_{ij}(x_j - \mu_j) + \eta(0, 1/\tau_i)
\end{aligned}
\end{equation}
\end{small}

where network parameters $\mu_i, \tau_i$, and $b_{ij}$ were chosen as independent realizations of Uniform([0.1, 1]).  Afterward, we computed the scores of all possible DAGs for the simulated data and calculated the rank of the generating DAG among all other DAGs. The simulation was repeated for 500 iterations.  Figure~\ref{Fig1} shows the average rank of the generating network structure for different scoring metrics as a function of the sample size.

As is shown in Figure~\ref{Fig1}, the convergence rate of the rank is the fastest for the RNML metric. The MDL metric and the BIC metric show similar performance while the AIC metric has the lowest convergence rate.

\begin{figure}[ht]
\begin{center}
\centerline{\includegraphics[width=2.5in]{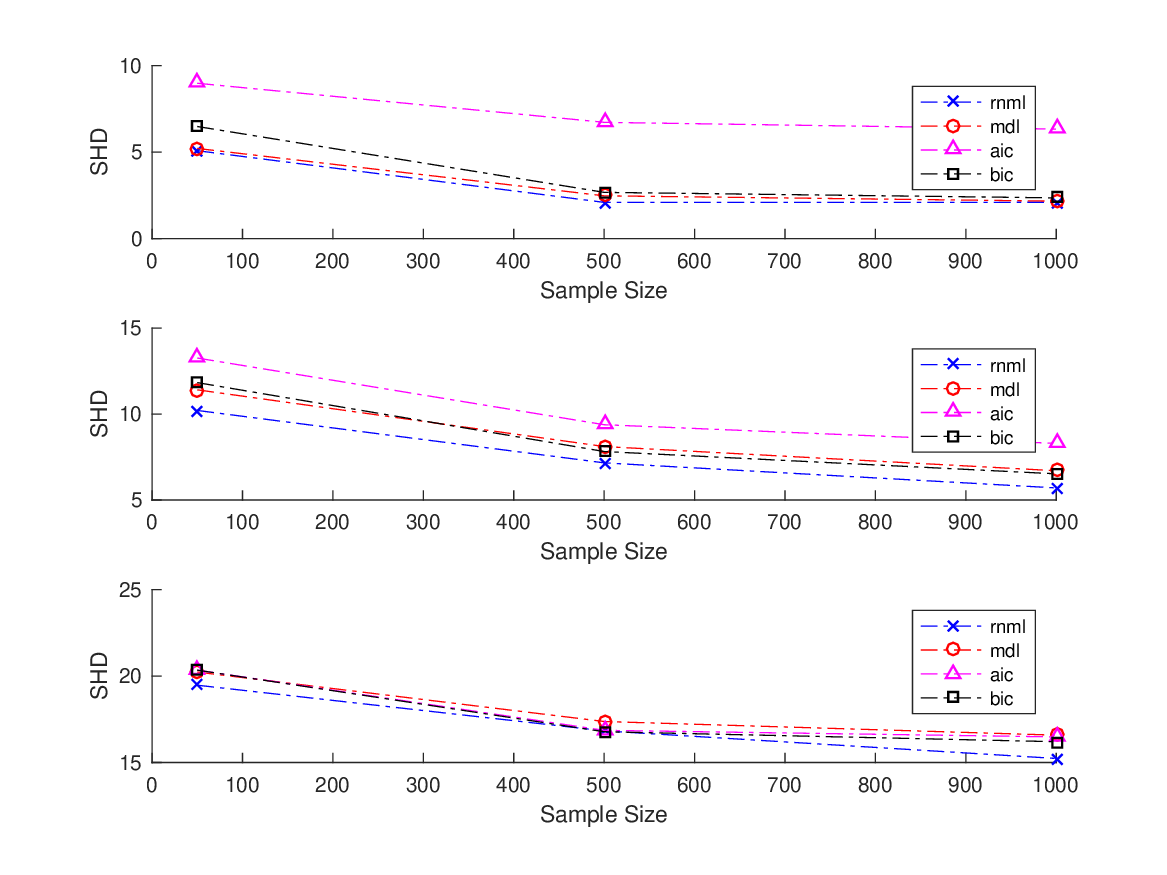}}
\caption{ Average SHD between the generating DAG and the prime DAG in simulations with $m=8$. The upper plot shows the results for graphs where the expected number of neighbours for each node was set to $nn = 2$. The middle and the lower plots show the results for $nn = 4$ and $nn = 6$. }
\label{Fig3.}
\end{center}
\end{figure}

\begin{figure}[ht]
\begin{center}
\centerline{\includegraphics[width=2.5in]{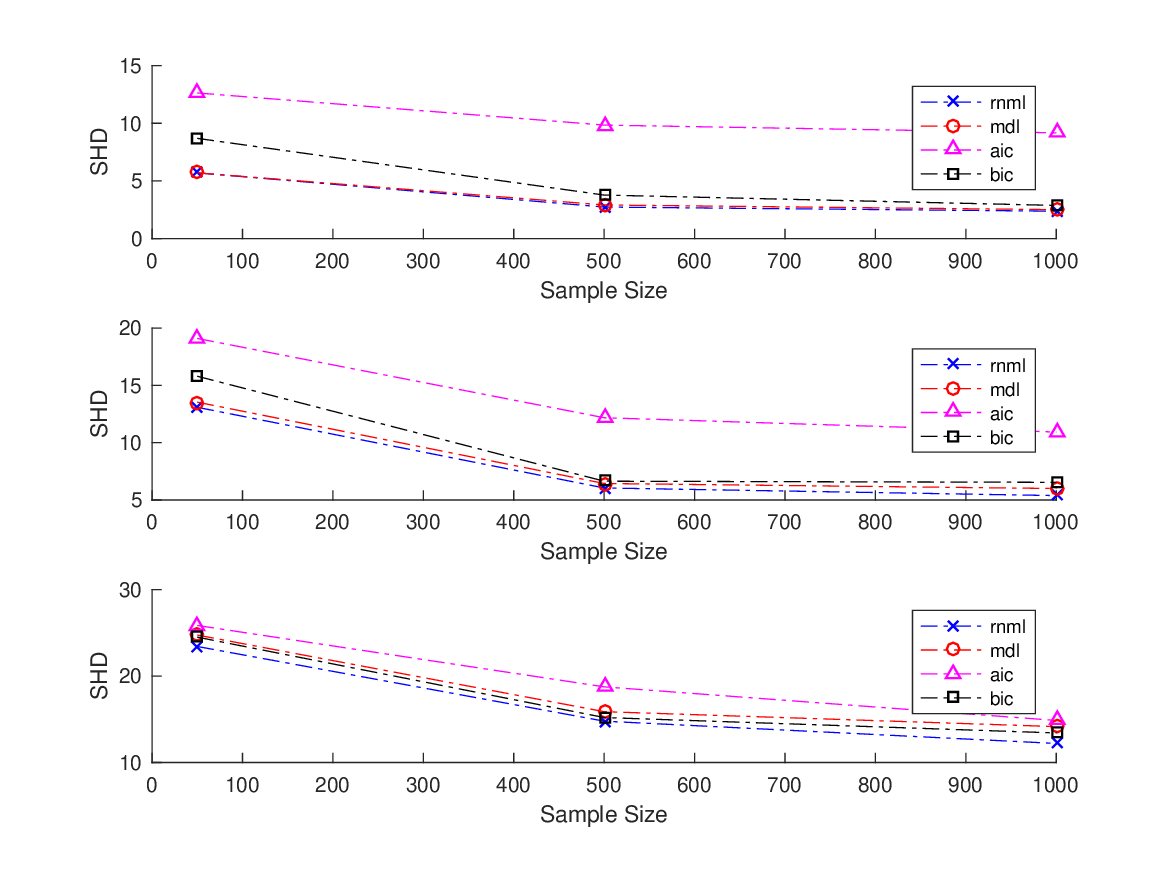}}
\caption{ Average SHD between the generating DAG and the prime DAG in simulations with $m=10$. The upper plot shows the results for graphs where the expected number of neighbours for each node was set to $nn = 2$. The middle and the lower plots show the results for $nn = 4$ and $nn = 6$. }
\label{Fig4.}
\end{center}
\end{figure}

\begin{figure}[ht]
\begin{center}
\centerline{\includegraphics[width=2.5in]{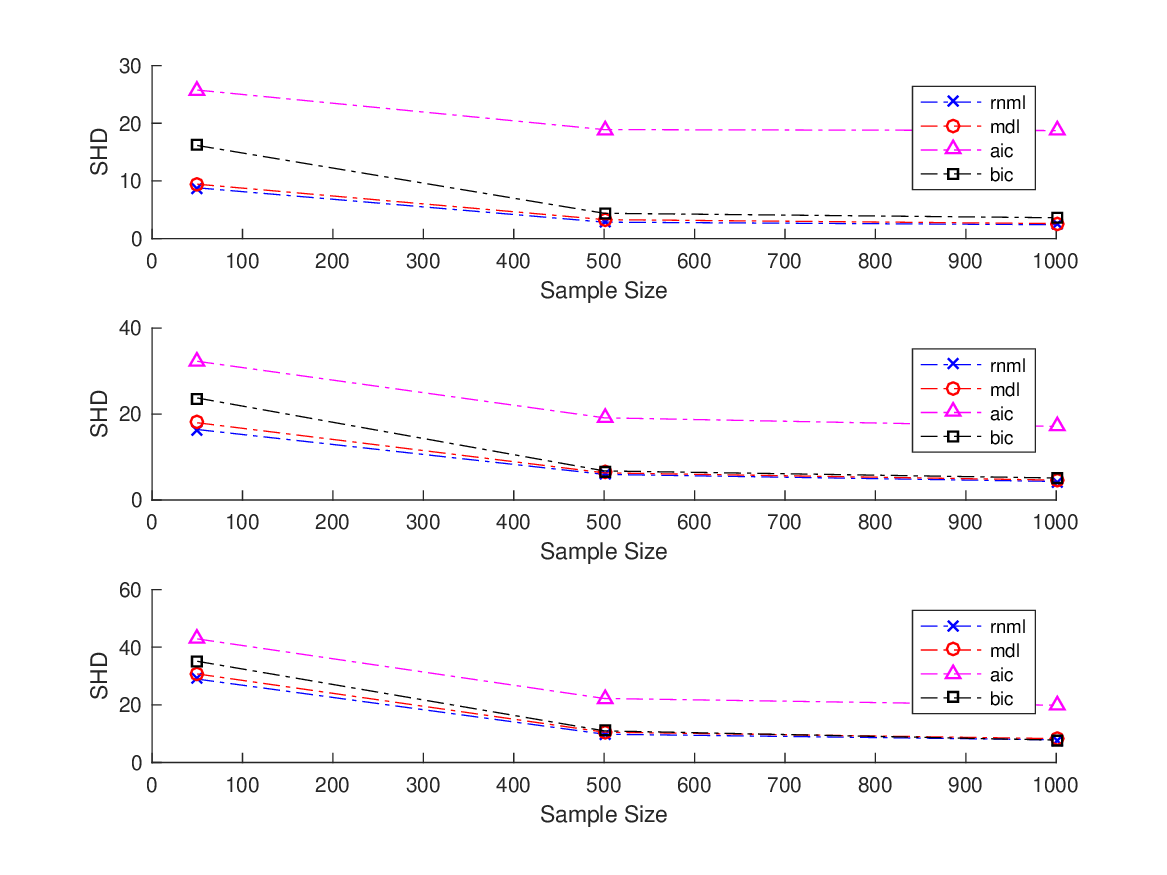}}
\caption{ Average SHD between the generating DAG and the prime DAG in simulations with $m=15$. The upper plot shows the results for graphs where the expected number of neighbours for each node was set to $nn = 2$. The middle and the lower plots show the results for $nn = 4$ and $nn = 6$. }
\label{Fig5.}
\end{center}
\end{figure}

\section{Performance For Larger Networks}
In this section, we analyze the performance of the four metrics against the sparsity of the generating graph for graphs having $m$ nodes, $m \in \{8, 10, 15\}$. We compare the performance of these metrics by evaluating the SHD of the prime DAG to the generating DAG.  

Our simulations start by selecting a random graph having a specified sparsity. Similar to the work of Kalisch et~al., we simulated graphs of different sparsity by controlling the expected number of connections, neighbours, of each node, $nn \in \{2, 4, 6\}$ \cite{kalisch2007estimating}. Thus, we extended the scope of our simulations beyond Erdos-Renyi (uniformly random) networks and examined the performance of the metrics for networks of varying sparsity. Afterward, a dataset of sample size $N$, $N \in \{50, 500, 1000\}$, was simulated based on the selected generating graph similar to the previous section.  To find the prime DAG, we used the optimal dynamic programming method of Silander et~al. \cite{silander2012simple}. In selecting the maximum parent size parameter in the implementation of this algorithm, we chose it equal to the maximum parent size of the generating graph itself. This way, the generating graph would be included in the possible solution set of our algorithm. For every of the 27 combinations of the control parameters (the control parameters being the number of nodes, expected number of neighbours, and sample size) we repeated the simulation for 100 iterations, each time for a randomly selected generating DAG. We then computed the average SHD between the generating and the prime DAG across these 100 iterations. Results are shown in Figures~2-4. 

Figures~2-4 show that the RNML metric consistently outperforms the other metrics irrespective of the size and the sparsity of the generating network. The proposed three-part MDL metric comes second in terms of performance, surpassing the AIC and the BIC metric. On average, aggregated across all sample size, sparsity, and network size values, the RNML metric outperformed the MDL metric by an SHD value of $0.6356$ and the other two metrics by an SHD value of $1.8704$. The average reduction in the SHD value for the RNML metric as compared to MDL and the best (smallest) of AIC/BIC is shown in Table~1 as a function of sample size.

\begin{table}[!t]
\renewcommand{\arraystretch}{1.3}
\caption{The average difference between the SHD of the prime graph to the generating graph when using RNML metric compared to that of MDL and the best (smallest) of AIC/BIC.}
\label{table_example}
\centering
\begin{tabular}{|c||c||c|}
\hline
Sample Size & MDL & BIC/AIC \\
\hline
\hline
50 & 0.7333 & 3.5900\\
\hline
500 & 0.6078 & 1.4422 \\
\hline
1000 & 0.5656 & 0.5789\\
\hline
\end{tabular}
\end{table}

\section{Conclusion}
In this paper, we introduced two new scoring metrics based on the MDL principle for Gaussian networks. These scores are asymptotically consistent, have simple to calculate closed-form expressions, and are parameter-free. They are furthermore decomposable and therefore compatible with the most Bayesian network search procedures. Our evaluation of the proposed metrics suggests that the proposed metrics have better performance than the AIC and the BIC metrics. The proposed RNML metric specially outperforms all other metrics consistently and regardless of the size and the sparsity of the generating graph and the sample size (see Figure~1 and Figures~2-4).

The RNML metric proposed here can be thought of as a continuous domain extension to the FNML metric proposed for discrete Bayesian networks \cite{silander2008factorized}. However, unlike the FNML metric, the RNML metric was not tuned to be decomposable, rather, decomposability came naturally to our formulation. More specifically, a major theoretical contribution of ours in this paper was to show that a global RNML formulation for a Gaussian network is equivalent to applying a RNML model selection criterion at each local distribution. 

\begin{quote}
\begin{scriptsize}
\bibliography{/Users/Borzou/Documents/Research/My_Publications/Arxiv_RNML/Arxiv_RNML_ref}
\bibliographystyle{aaai}
\end{scriptsize}
\end{quote}

\end{document}